\documentclass{article}

\usepackage{arxiv}

\usepackage[utf8]{inputenc} 
\usepackage[T1]{fontenc}   

\usepackage{hyperref}      
\hypersetup{hidelinks}

\usepackage{url}            
\usepackage{booktabs}       
\usepackage{amsfonts}  
\usepackage{amsmath}
\usepackage{nicefrac}       
\usepackage{microtype}      
\usepackage{cleveref}   
\usepackage{graphicx}
\usepackage{float}
\usepackage[numbers]{natbib}
\usepackage{doi}

\usepackage{caption}
\usepackage{titlesec}
\renewcommand{\headeright}{}

\usepackage{algorithm}
\usepackage{algpseudocode} 

\title{Architecting AgentOS: From Token-Level Context to Emergent System-Level Intelligence}

\author{
ChengYou~LI$^{1}$,
XiaoDong~Liu$^{2}$,
XiangBao~Meng$^{1}$,
XinYu~Zhao$^{3}$\\[4pt]
$^{1}$Yishu Research\\
$^{2}$Fukuoka Institute of Technology, Japan\\
$^{3}$National University of Singapore
}

\begin{document}

\maketitle

\begin{abstract}
	The paradigm of Large Language Models (LLMs) is undergoing a fundamental transition from static inference engines to dynamic, autonomous cognitive systems. While current research primarily focuses on scaling context windows or optimizing prompt engineering, the theoretical bridge between micro-scale token processing and macro-scale systemic intelligence remains fragmented. This paper proposes AgentOS, a holistic conceptual framework that redefines the LLM as a "Reasoning Kernel" governed by structured operating system logic. \\Central to this architecture is Deep Context Management, which conceptualizes the context window as an Addressable Semantic Space rather than a passive buffer. We systematically deconstruct the transition from discrete sequences to coherent cognitive states, introducing mechanisms for Semantic Slicing and Temporal Alignment to mitigate cognitive drift in multi-agent orchestration. By mapping classical OS abstractions—such as memory paging, interrupt handling, and process scheduling—onto LLM-native constructs, this review provides a rigorous roadmap for architecting resilient, scalable, and self-evolving cognitive environments. Our analysis asserts that the next frontier of AGI development lies in the architectural efficiency of system-level coordination.
\end{abstract}


\section{Introduction}
\subsection{The Post-Von Neumann Paradigm Shift}
For over half a century, the Von Neumann architecture has dominated the landscape of computing, characterized by the rigid separation of the Central Processing Unit (CPU) and memory. In this classical framework, data are inert and logic is deterministic. The emergence of Large Language Models (LLMs), however, has introduced a radical "Reasoning Kernel" (RK) where memory (Context) functions as an active computational substrate. Unlike traditional processors that execute bitwise arithmetic, the RK performs Contextual Transformations, synthesizing information and simulating cognitive reasoning through parallel self-attention(see Fig. 1.1).
\begin{figure}[H]
\centering
\includegraphics[width=\linewidth]{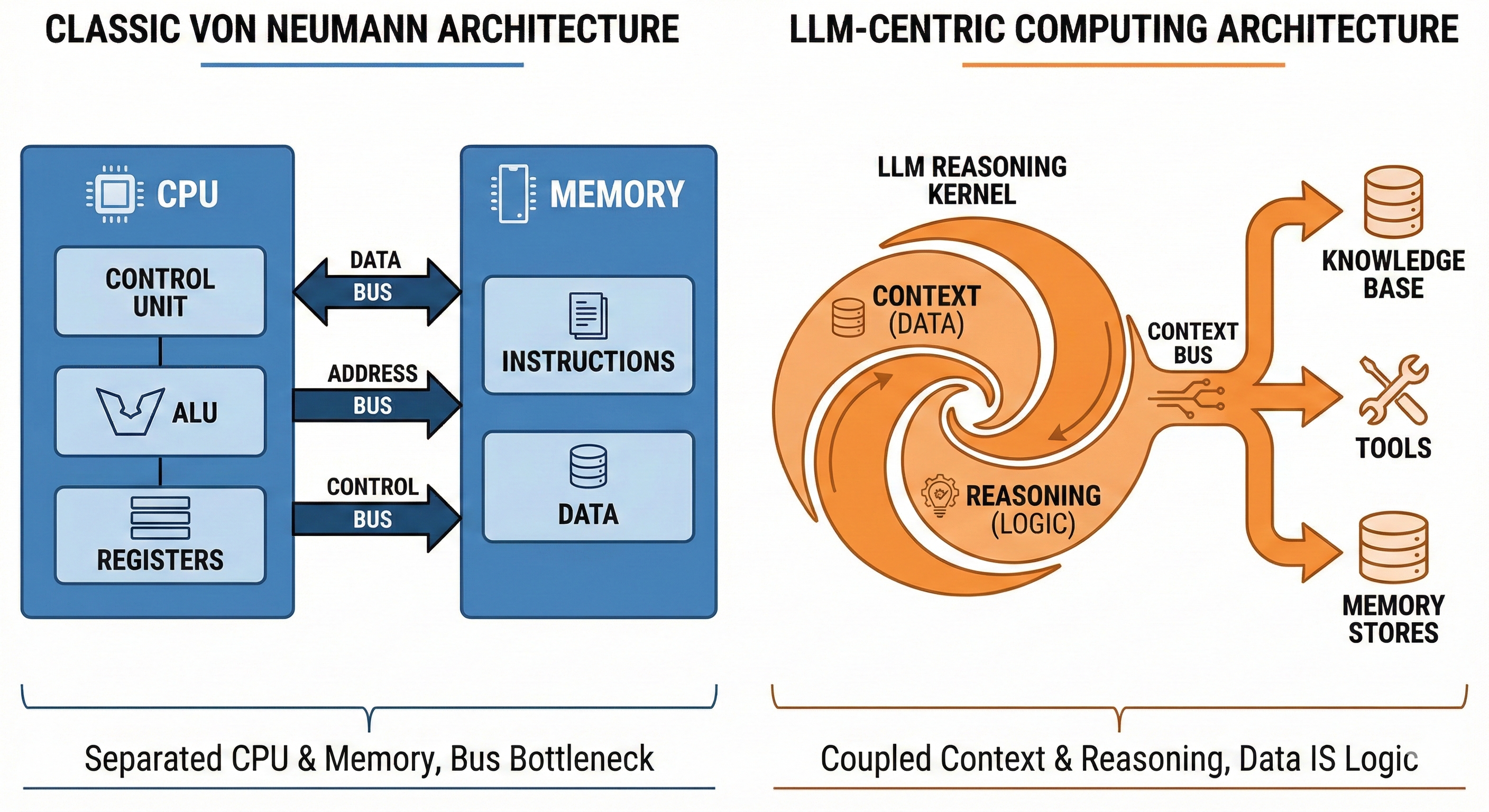}
\caption{Fig.~1.1}
\label{fig:post-von-neumann}
\end{figure}

\subsection{The Architectural Gap and Contextual Volatility}
Despite the impressive capabilities of LLMs, current agentic deployments suffer from what we define as the "Architectural Gap." Most contemporary frameworks (e.g.AutoGen, BabyAGI) treat the LLM as a stateless API. This "Model-as-a-Service" mentality overlooks the systemic requirements of long-form, autonomous reasoning. \\Specifically, agents encounter Spatio-temporal Dissociation:
\begin{itemize}
\item Spatial Decay: In long-context tasks, information becomes diluted, leading to the "lost-in-the-middle" phenomenon, where the RK fails to attend to critical semantic clusters.
\item Temporal Drift: In asynchronous multi-agent orchestration, independent reasoning threads diverge over time, resulting in a loss of collective "State-of-Truth."
\end{itemize}
Traditional approaches to scaling the context window offer a larger "canvas" but do not provide the "indexing" or "synchronization" necessary for high-fidelity system-level intelligence.
\subsection{Related Work and Gap Analysis}
The quest for a "Systemic LLM" has led to early explorations such as MemGPT, which introduced hierarchical memory management, and AIOS, which proposed a basic kernel for process scheduling. While pioneering, these efforts primarily address Application-level management. They lack a formal theory of how Discrete Tokens—the physical layer—evolve into Emergent Intelligence—the systemic layer.
\\We identify three critical deficiencies in current literature(see Fig. 1.3):
\begin{itemize}
\item The Granularity Problem: Existing systems treat the context window as a monolithic block of tokens rather than an addressable set of semantic units.
\item The Synchronization Problem: There is no standardized protocol for "cognitive de-confliction" when multiple agents access a shared context space.
\item The Resource Problem: There is a lack of formalisms for "Cognitive Bandwidth" and the overhead of context switching between disparate reasoning tasks.
\end{itemize}
\begin{figure}[!t]
\centering
\includegraphics[width=\linewidth]{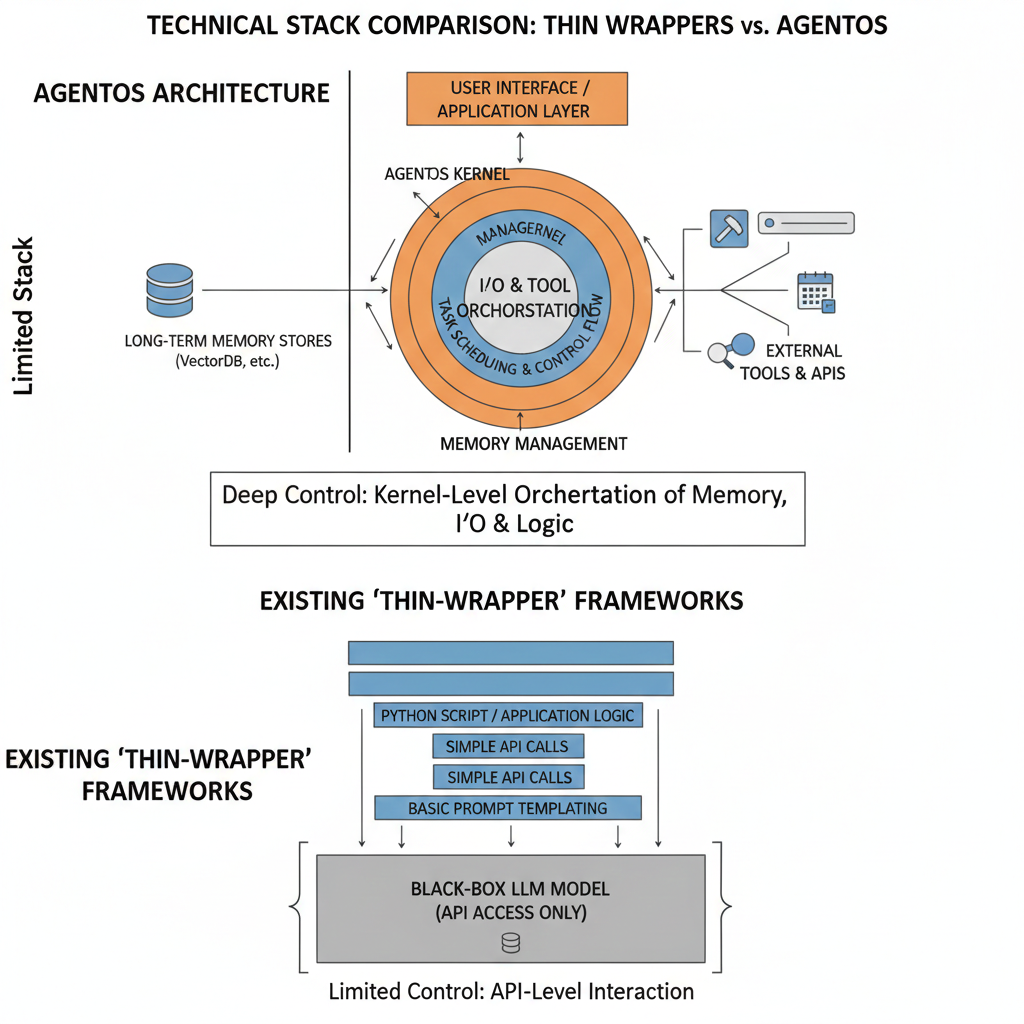}
\caption{Fig.~1.3}
\label{fig:post-von-neumann}
\end{figure}

\subsection{Contributions of this Review}
To address these challenges, this paper proposes AgentOS as a comprehensive system abstraction. Our contributions are three-fold:
\begin{itemize}
\item Functional Abstraction: We map classical OS concepts (Process, Memory Hierarchy, I/O) to LLM-native cognitive constructs.
\item Theory of Semantic Slicing: We provide a micro-mechanical analysis of how tokens aggregate into addressable semantic units.
\item Synchronization Framework: We propose a roadmap for Perception Alignment, ensuring temporal consistency across distributed agent ecosystems.
\end{itemize}

\section{The Anatomy of AgentOS: Systemic Abstractions}
To bridge the gap between stochastic token sequences and deterministic system behavior, AgentOS introduces a layered abstraction that treats cognitive processes as manageable system resources. This section deconstructs the architectural pillars of AgentOS.
\subsection{The Reasoning Kernel (RK) and Logic State Management}
The Reasoning Kernel (RK) is the central processing unit of AgentOS. Unlike traditional CPUs that operate on a fixed Instruction Set Architecture (ISA), the RK operates on a Contextual Transition Function. We define the RK's operation as a mapping:
\begin{equation}
\mathcal{F} : (S_t, C_{addr}) \rightarrow S_{t+1}
\end{equation}
where $S_t$ is the current cognitive state, and $\mathcal{C}_{addr}$ is the Addressable Context Space.
The OS layer provides a Reasoning Control Block (RCB)—analogous to a Process Control Block (PCB) in Unix—to track the state of each reasoning thread, including its attention focus, active tool-calls, and semantic stack depth. This ensures that the RK remains "context-aware" across asynchronous task switching.
\subsection{Cognitive Memory Hierarchy (CMH) and S-MMU}
A major limitation of vanilla LLMs is the flat structure of the context window. AgentOS implements a Cognitive Memory Hierarchy (CMH) managed by the Semantic Memory Management Unit (S-MMU). The S-MMU is responsible for Semantic Paging—the process of loading and unloading context slices based on their task-relevance(see Fig. 2.2).
\begin{itemize}
\item L1 Cache (Immediate Attention): This is the active KV-Cache of the Transformer. It has the lowest latency but is limited by the $O(n^2)$ complexity of the self-attention mechanism.
\item L2 RAM (Deep Context): Managed as an Addressable Semantic Space. The S-MMU utilizes a Semantic Page Table (SPT) to track "Semantic Slices." When a reasoning thread shifts focus, the S-MMU performs a "Swap-out" of irrelevant slices to L2, maintaining only the core logical anchors in L1.
\item L3 Storage (Knowledge Base): This comprises external Vector Databases and RAG systems. Data in L3 is "cold" and requires an explicit I/O request to be paged into the S-MMU for active reasoning.
\end{itemize}
\begin{figure}[H]
\centering
\includegraphics[width=\linewidth]{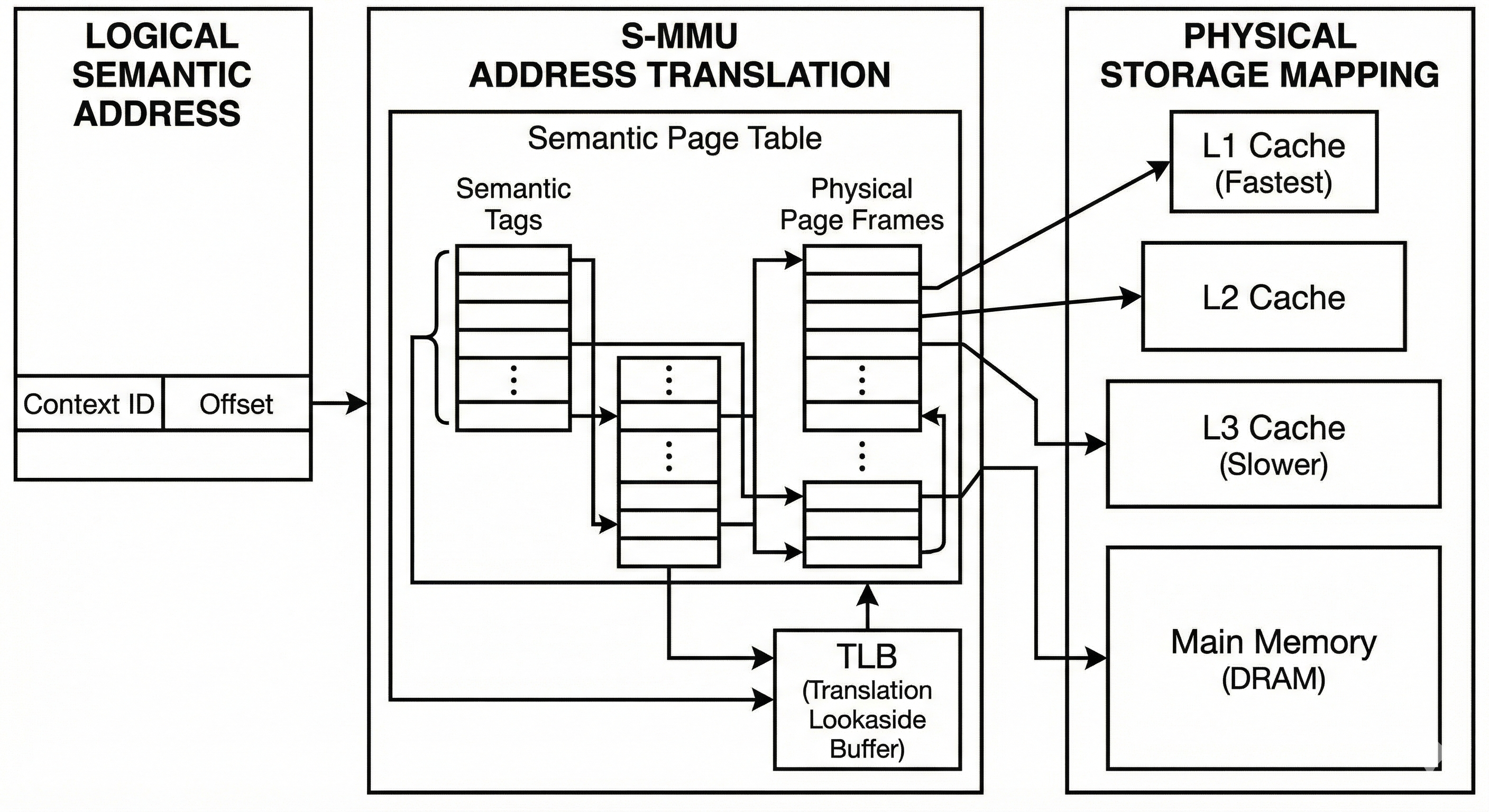}
\caption{Fig.~2.2}
\label{fig:post-von-neumann}
\end{figure}
\subsection{The I/O Subsystem and Reasoning Interrupts}
In AgentOS, external tools (e.g., Python interpreters, Search APIs) are treated as Peripheral Devices. Interaction with these devices is governed by the Reasoning Interrupt Cycle (RIC).
\\When the RK generates a tool-invocation sequence, the OS triggers a Reasoning Interrupt. The system saves the current semantic state, executes the tool call, and performs Perception Alignment—a process of filtering and re-coding the tool's output to ensure it fits the current semantic schema of the context window.

\begin{algorithm}[H]
\caption{Semantic Context Switching and Interrupt Handling}
\label{alg:interrupt}
\begin{algorithmic}[1]
\State \textbf{Initialize} Reasoning Thread $T_i$, Active Context $\mathcal{C}_{L1}$
\While{$T_i$ is not terminated}
  \State $Token_{next} \gets RK.Execute(\mathcal{C}_{L1})$
  \If{$Token_{next}$ indicates TOOL\_REQUEST}
    \State \textbf{SIGNAL} Reasoning Interrupt
    \State \textbf{STORE} Active Slice $\sigma_{curr} \to L2\_Memory$
    \State $Raw\_Data \gets External\_Device.Call(Params)$
    \State $\sigma_{aligned} \gets Perception\_Align(Raw\_Data)$
    \State \textbf{RELOAD} $\sigma_{curr}$ and \textbf{APPEND} $\sigma_{aligned} \to \mathcal{C}_{L1}$
  \EndIf
\EndWhile
\end{algorithmic}
\end{algorithm}

\subsection{The Cognitive Scheduler: Managing Bandwidth}
The Cognitive Scheduler is responsible for allocating RK cycles to multiple competing agents. Unlike traditional schedulers that optimize for CPU time, the AgentOS scheduler optimizes for Cognitive Fidelity and Token Efficiency. It utilizes a Priority-based Semantic Scheduling algorithm, ensuring that high-stakes reasoning threads (e.g., safety monitoring) receive preferential access to the RK's attention.
\section{The Micro-Mechanics: From Sparse Tokens to Contextual Clusters}
The foundational layer of AgentOS rests upon the transition from stochastic token prediction to deterministic state management. This section provides a micro-scale analysis of how self-attention mechanisms facilitate the emergence of "Semantic Slices," the atomic units of the S-MMU.
\subsection{Attention as a Structural Filter}
In the Transformer architecture, the self-attention mechanism computes a weighted representation of the input sequence. Each attention head $h$ in layer $L$ identifies specific relational dependencies. We hypothesize that these dependencies are not uniformly distributed but gravitate toward Semantic Anchors—tokens that hold disproportionate cognitive weight (e.g., entity definitions, logical operators, or temporal markers).
\\We define the Contextual Information Density (CID) at position $t$ as the negative entropy of the attention distribution across all heads $H$: 
\begin{equation}
D(t)=1-\left[
-\frac{1}{H}
\sum_{i=1}^{H}
\sum_{j=1}^{t}
\alpha_{i,j}\log(\alpha_{i,j})
\right]
\end{equation}
where $\alpha_{i,j}$ represents the attention weight between token $i$ and token $j$. A sharp gradient in $D(t)$ indicates a transition in the reasoning flow, marking a potential Semantic Boundary.

\subsection{The Formation of Semantic Slices}
Unlike traditional LLM inference, which treats the context window as a monolithic, sliding buffer of $N$ tokens, AgentOS implements Dynamic Semantic Slicing. This process aggregates tokens into coherent clusters $\{\sigma_1, \sigma_2, ..., \sigma_k\}$ based on their mutual information and attention cohesion.
\\These slices are functionally equivalent to "Cognitive Pages". When $D(t)$ falls below a stability threshold $\epsilon$, the current sequence is "finalized" into a slice and assigned a Semantic Hash. This hash allows the OS to perform rapid indexing and deduplication within the L2 Semantic RAM, effectively solving the "redundant context" problem that plagues multi-agent dialogues(see Fig. 3.2).

\begin{figure}[H]
\centering
\includegraphics[width=\linewidth]{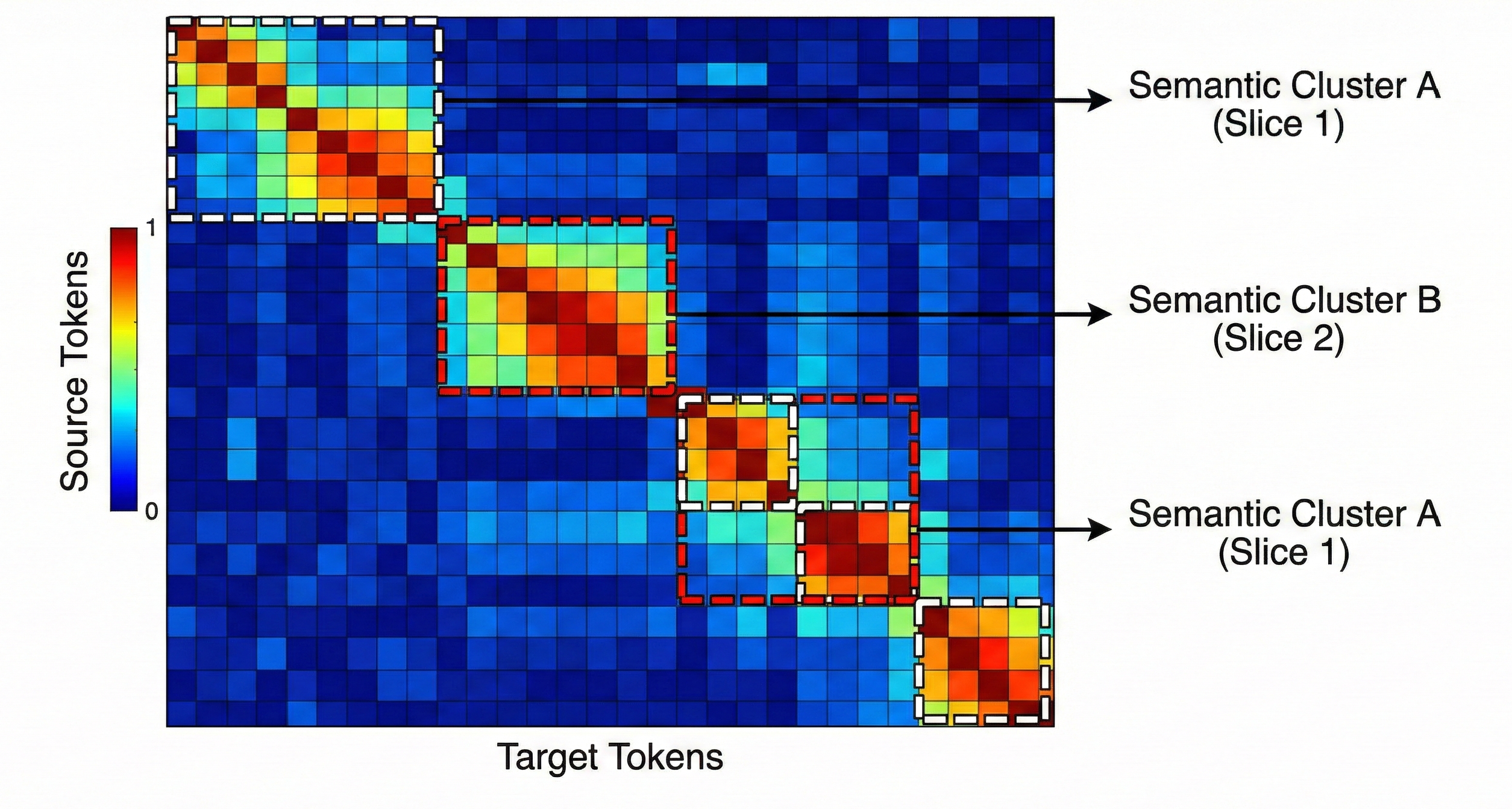}
\caption{\textbf{Fig.~3.2} Attention Matrix Heatmap revealing emerging Block Structures. These blocks visually demonstrate the natural aggregation of tokens into Semantic Slices.}
\label{fig:attention-heatmap}
\end{figure}

\subsection{The Transition from Token to State}
The "Genesis" of intelligence within AgentOS occurs when these slices are transformed into Systemic States. In classical LLMs, the hidden state $h_t$ is volatile. In AgentOS, the OS layer performs a State Compression on each slice, distilling the raw tokens into a persistent Latent Schema(see Fig. 3.3).
\\This schema serves as the interface for the Reasoning Kernel (RK). By operating on schemas rather than raw tokens, AgentOS achieves:
\begin{itemize}
\item Linear Scalability: The RK only attends to the most relevant schemas, bypassing the $O(n^2)$ limitation for distant, irrelevant tokens.
\item Deterministic Retrieval: The S-MMU can fetch precise semantic states from L2/L3 using the Semantic Page Table (SPT), ensuring that "Perception Alignment" is grounded in historical truth rather than probabilistic hallucination.
\end{itemize}

\begin{figure}[H]
\centering
\includegraphics[width=\linewidth]{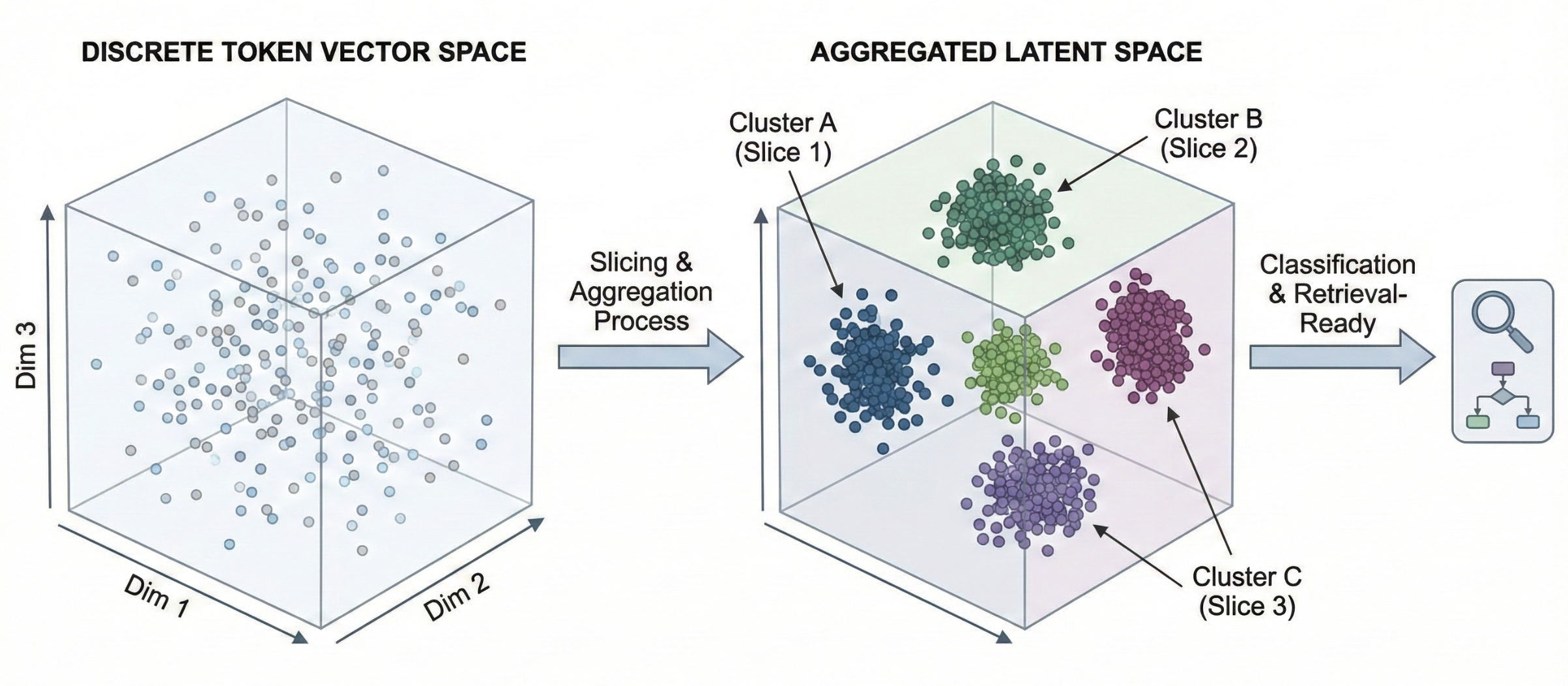}
\caption{\textbf{Fig.~3.3} Evolution from discrete to aggregated latent space. The transition from sparse token vectors to clustered, semantically organized slices in the latent space, facilitating efficient retrieval and classification.}
\label{fig:latent-evolution}
\end{figure}

\section{Emergence via Synchronization: The Multi-Agent Orchestration}
When multiple AgentOS instances interact within a shared task environment, the primary challenge shifts from individual reasoning to Collective Coherence. In an asynchronous multi-agent ecosystem, agents often operate on disparate temporal scales and divergent semantic states, leading to a phenomenon we term Cognitive Entropy.
\subsection{The Challenge of Asynchronous Cognitive Drift}
In classical multi-agent systems, communication is typically turn-based and sequential. However, in a true AgentOS environment, agents execute tasks concurrently, interacting with various external tools and internal L2/L3 memory layers. This asynchrony introduces Cognitive Drift ($\Delta \psi$). We define this drift as the divergence between an agent's local perception of the environment and the objective "State-of-Truth" ($\mathcal{S}_{global}$):
\begin{equation}
\Delta \psi_i(t)=
\int_{0}^{t}
\left\lVert
\nabla \Phi_i(\sigma,\tau)-\nabla S_{\text{global}}(\tau)
\right\rVert
\, d\tau
\end{equation}
As $\Delta \psi$ accumulates, agents begin to generate conflicting "Semantic Slices," leading to logical deadlocks or hallucinatory contentions. To prevent this, AgentOS must implement a system-level synchronization protocol(see Fig. 4.1).

\begin{figure}[H]
\centering
\includegraphics[width=\linewidth]{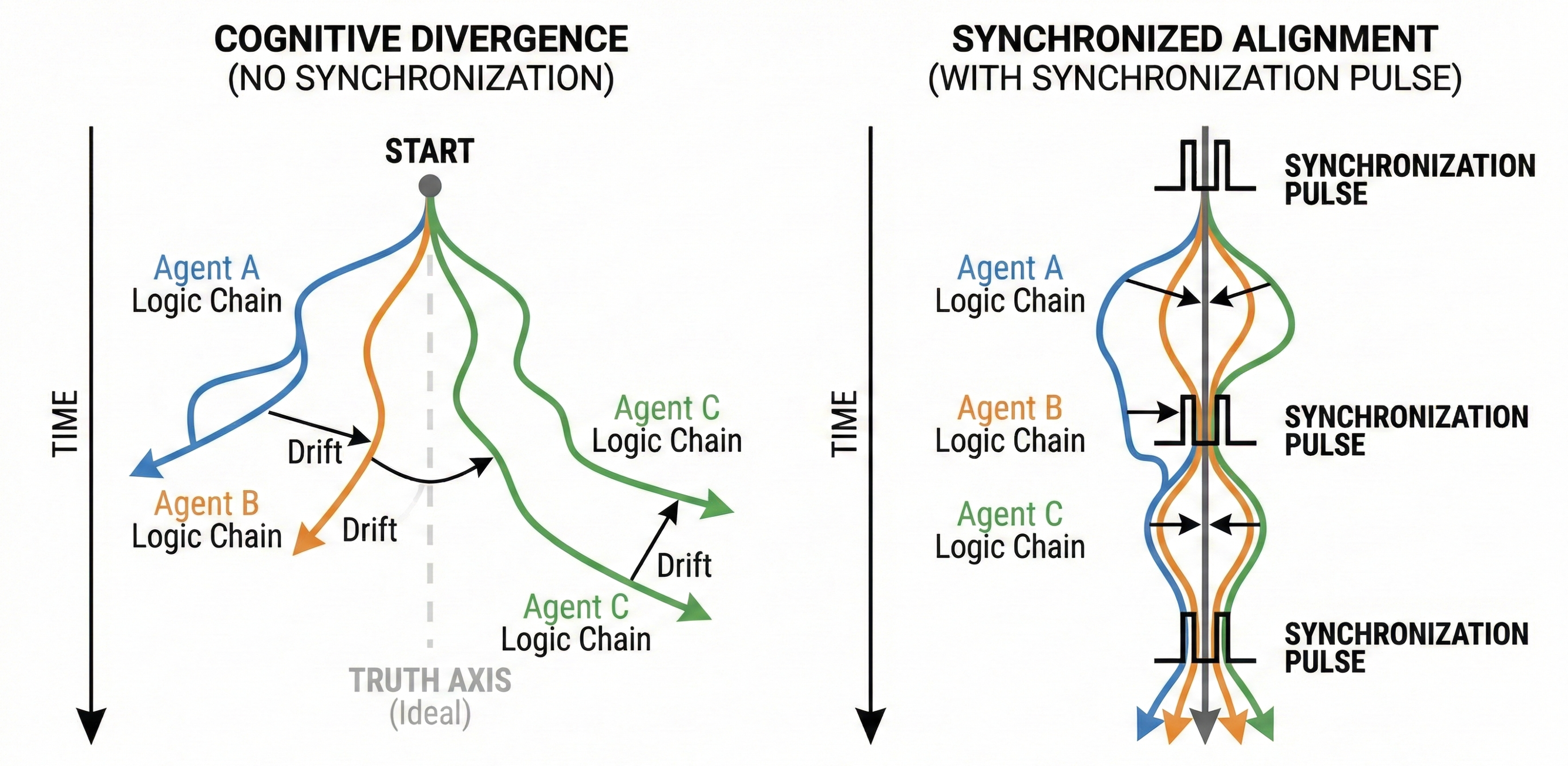}
\caption{\textbf{Fig.~4.1} Comparison of multi-agent logic chains over time, showing divergence without synchronization (left) and alignment with periodic synchronization pulses (right).}
\label{fig:multi-agent-sync}
\end{figure}

\subsection{Cognitive Sync Pulses (CSP) and Logical Time}
To mitigate drift, AgentOS introduces Cognitive Sync Pulses (CSP). Unlike the constant-frequency clock of a traditional CPU, a CSP is an Event-Driven Interrupt triggered by the S-MMU when a significant semantic transition is detected (e.g., the completion of a tool-call or the formation of a critical logical anchor).
During a CSP, the OS orchestrates a Contextual Checkpoint. The Reasoning Kernel pauses active threads to perform a Global State Reconciliation. This ensures that all participating agents are "cognitively paged" into the same version of the addressable semantic space, effectively acting as a "distributed shared memory" for intelligence.
\subsection{Perception Alignment: The Gateway to Collective Emergence}
The ultimate goal of AgentOS is to facilitate Emergent Intelligence, where the collective output exceeds the sum of individual LLM capacities. This is achieved through the Perception Alignment Protocol.
\\When agents synchronize, they do not simply exchange all tokens—which would be computationally prohibitive. Instead, they perform Advantageous Timing Alignment. By identifying "High-Confidence Windows" within the reasoning flow, the system selects the optimal moments to merge disparate semantic slices. This mechanism ensures that only the most "logically robust" information is propagated through the system, filtering out the noise inherent in probabilistic inference(see Fig. 4.3).

\begin{figure}[H]
\centering
\includegraphics[width=\linewidth]{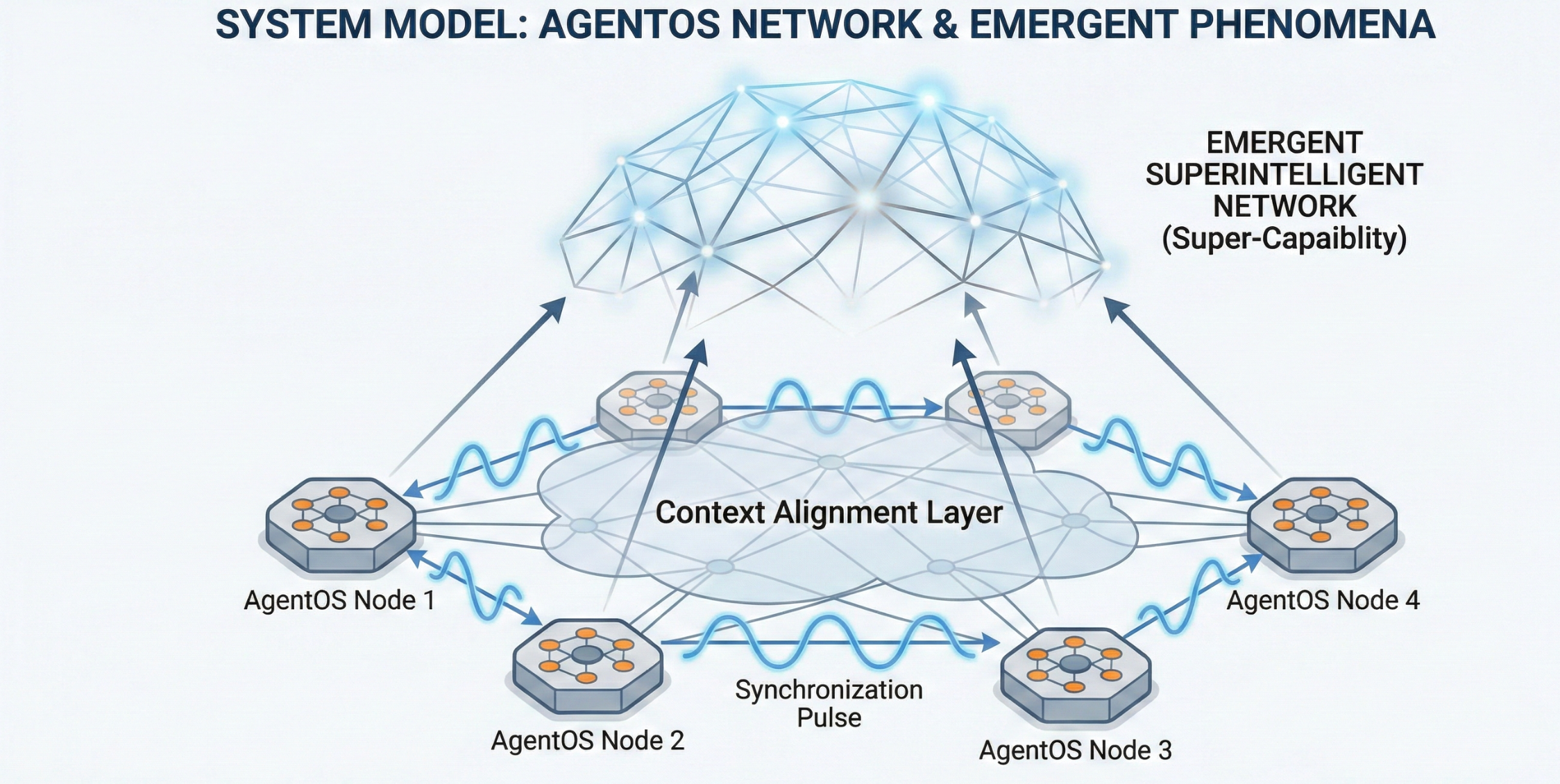}
\caption{\textbf{Fig.~4.3} A distributed AgentOS network where synchronization pulses through a context alignment layer foster emergent superintelligent capabilities beyond individual nodes.}
\label{fig:agentos-network}
\end{figure}

\paragraph{Theoretical Implementation Note:}
The specific algorithmic realization of this alignment—detecting the precise temporal slices for optimal matching—represents a critical sub-domain of AgentOS research, which we define as the "Advantageous-Timing Matching Mechanism." This mechanism is essential for maintaining the stability of the Sync Stability Index ($\Gamma$) in complex, real-world deployments.
\subsection{The Emergence of Systemic Consciousness}
Through the interplay of Semantic Slicing and Perception Alignment, AgentOS moves from being a "tool for thought" to an "environment for intelligence." In this state, the multi-agent system exhibits Systemic Persistence, maintaining a coherent long-term goal trajectory despite the underlying volatility of individual token predictions.

\section{Systemic Evaluation and Theoretical Constraints}
To transition AgentOS from a conceptual framework to an engineering standard, we must establish a rigorous taxonomy of metrics and identify the fundamental constraints that govern its scalability.
\subsection{Metrics for Cognitive Operating Systems}
Traditional benchmarks like MMLU or HumanEval measure the raw intelligence of an LLM, but they fail to capture the architectural efficiency of an AgentOS. We propose the following system-level metrics(see Fig. 5.1):
\begin{itemize}
\item Cognitive Latency ($L_c$): The temporal overhead introduced by the OS layer, measured from the moment an external interrupt (I/O or CSP) occurs to the moment the Reasoning Kernel (RK) resumes a stable state transition.
\item Contextual Utilization Efficiency ($\eta$): Defined as the ratio of "Information-Gain Tokens" to "Total Processed Tokens." A high $\eta$ indicates that the S-MMU is effectively filtering noise and only paging high-value semantic slices into L1.
\item 	\begin{equation}
			\eta=\frac{\sum \mathrm{IG}(\sigma_{\text{active}})}
			{\sum \mathrm{Tokens}_{\text{processed}}}
			\end{equation}
\item Sync Stability Index ($\Gamma$): The probability that a multi-agent cluster maintains a unified state vector $\Delta \psi < \epsilon$ over a prolonged execution cycle. This measures the robustness of the Perception Alignment Protocol.
\end{itemize}

\begin{figure}[!t]
\centering
\includegraphics[width=\linewidth]{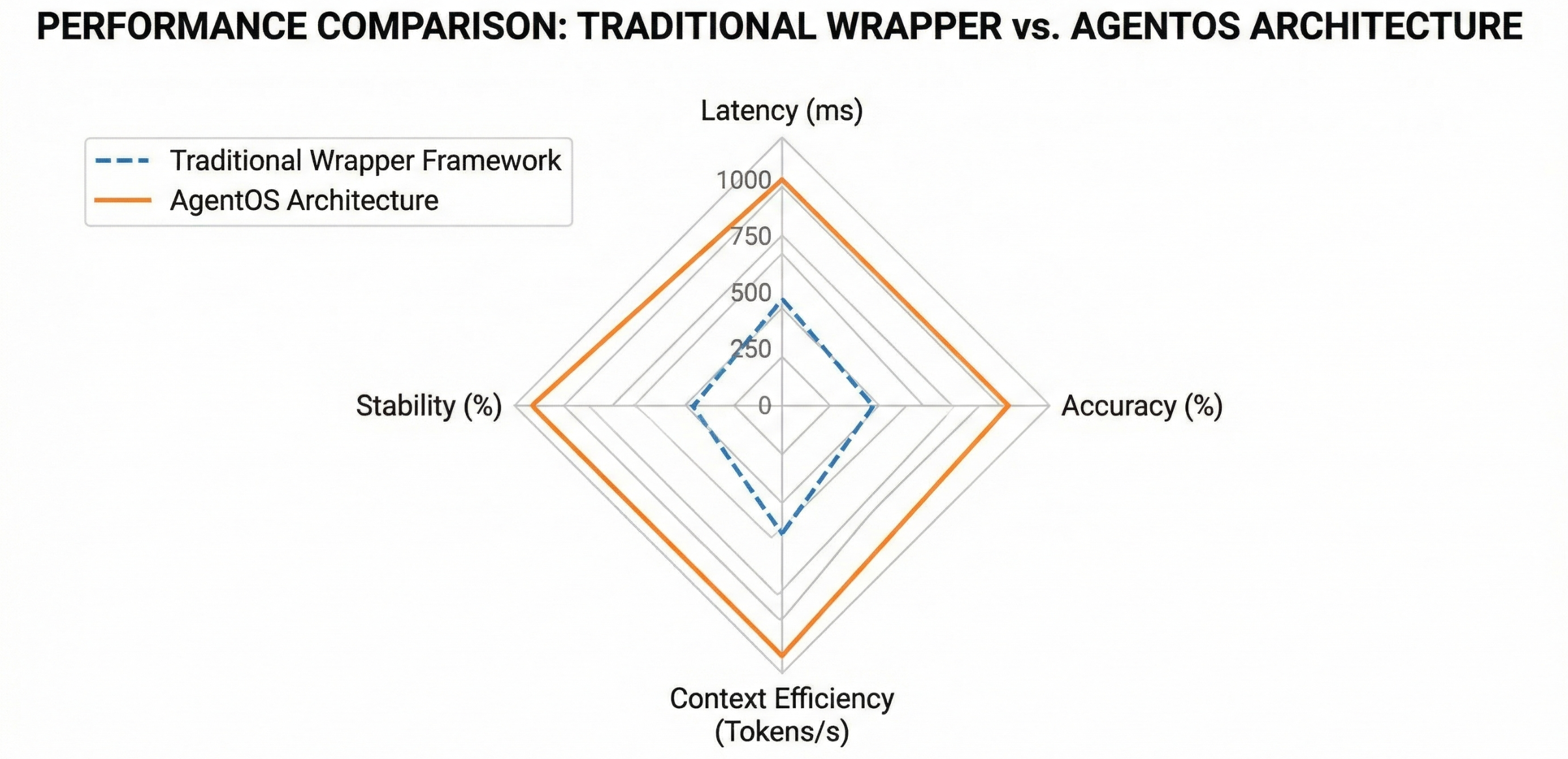}
\caption{\textbf{Fig.~5.1} Radar chart comparison demonstrating the system-level superiority of AgentOS across key metrics, particularly in accuracy, efficiency, and stability, compared to traditional wrapper-based approaches.}
\label{fig:agentos-radar}
\end{figure}

\subsection{Theoretical Constraints: The "Cognitive Bottleneck"}
Despite the advantages of AgentOS, it is subject to the Law of Diminishing Cognitive Returns. We identify three primary constraints(see Fig. 5.2):
\begin{itemize}
\item Context-Switching Penalty: Each time the RK switches between disparate reasoning threads, the S-MMU must perform a KV-Cache reload. As the number of concurrent threads $N$ increases, the system risks entering a state of "Cognitive Thrashing," where more cycles are spent on synchronization than on actual reasoning.
\item Semantic Paging Latency: The speed of the S-MMU is bound by the throughput of the L2/L3 memory interface. In massive-scale AgentOS deployments, the retrieval of historical "Semantic Pages" may become the primary bottleneck.
\item The Entropy Barrier: As the multi-agent system grows, the complexity of maintaining the State-of-Truth increases non-linearly. The cost of Perception Alignment follows $O(k^2)$ relative to the number of interacting agents $k$.
\end{itemize}

\begin{figure}[H]
\centering
\includegraphics[width=\linewidth]{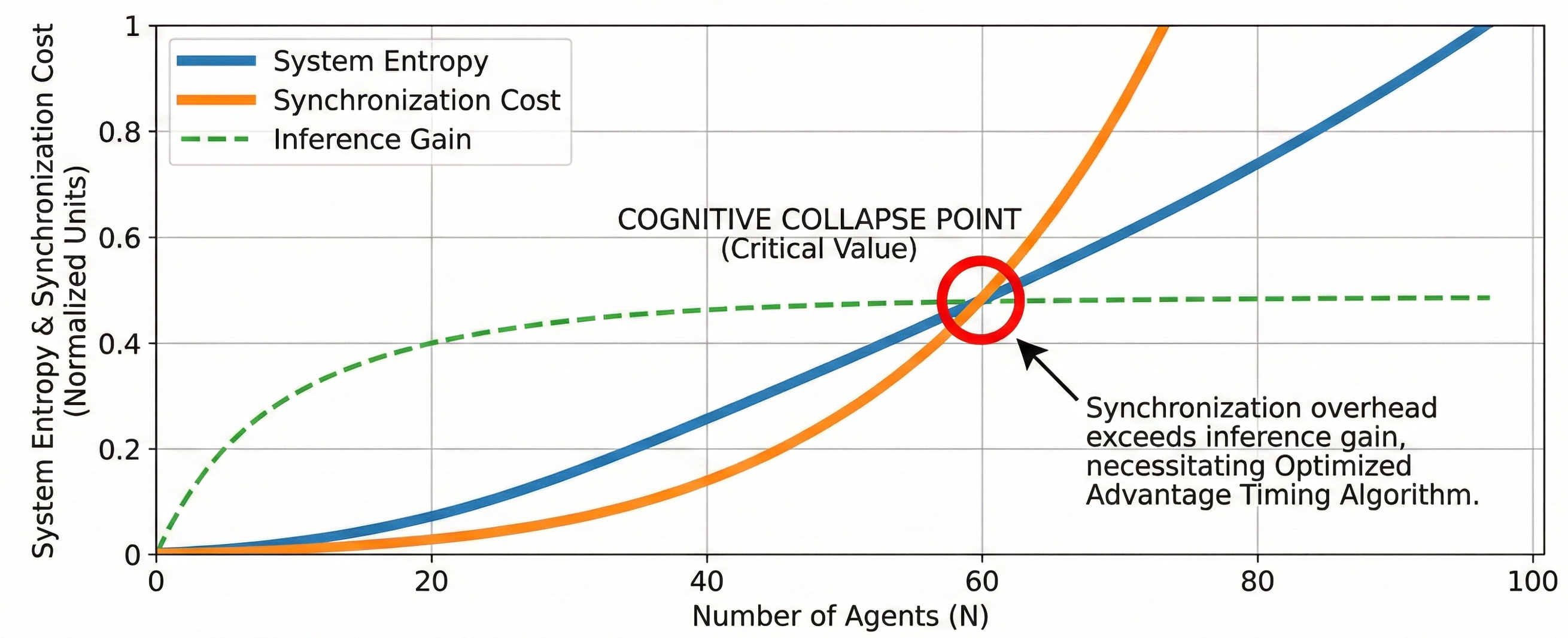}
\caption{\textbf{Fig.~5.2} Evolution of system entropy, synchronization cost, and inference gain with increasing agent count. The graph illustrates the escalating system entropy and synchronization cost as the number of agents increases. The ``Cognitive Collapse Point'' marks the critical threshold where synchronization overhead outweighs inference benefits, underscoring the necessity for the proposed optimized advantage timing algorithm.}
\label{fig:entropy-evolution}
\end{figure}

\section{Conclusion and Future Work}
The transition from "Model-as-a-Service" to AgentOS marks the maturation of artificial intelligence from a probabilistic predictor to a systematic entity. By architecting a framework that treats Deep Context as an addressable, slicable, and synchronizable memory space, we provide the infrastructure necessary for true Emergent Intelligence.
\\Our analysis demonstrates that the genesis of higher-order cognition is not merely a function of parameter scaling, but a result of Architectural Orchestration. The proposed AgentOS paradigm offers a path toward AGI that is resilient to the volatility of token-level inference.

The immediate next step in the AgentOS roadmap is the formalization of the "Advantageous-Timing Matching Mechanism." Future research must focus on optimizing the S-MMU algorithms to reduce context-switching overhead and exploring hardware-level acceleration for Semantic Paging. As we move toward a future of ubiquitous multi-agent ecosystems, the principles of AgentOS will serve as the cornerstone for the next generation of resilient and self-evolving artificial intelligence.

\newpage
\clearpage
\appendix

\begin{center}
\large\bfseries Appendix
\end{center}

\section{Mathematical Foundations and Formalism}
\subsection{Summary of Notations}
To ensure clarity across the multi-disciplinary abstractions of AgentOS, Table A1 categorizes the primary notations used in our formalisms.

\begin{table}[H]
\centering
\caption{Summary of Main Notations}
\label{tab:A1}
\begin{tabular}{p{0.22\linewidth} p{0.53\linewidth} p{0.18\linewidth}}
\toprule
\textbf{Symbol} & \textbf{Description} & \textbf{Domain} \\
\midrule
$\mathcal{K}$ & The Reasoning Kernel (RK), represented as a transformation function. & System Core \\

$\mathcal{C}_{L1},\ \mathcal{C}_{L2}$ & Level 1 (Attention window) and Level 2 (Semantic RAM) context spaces. & Memory \\

$\sigma_i$ & The $i$-th \textbf{Semantic Slice}, the atomic unit of the S-MMU. & Data Unit \\

$\mathbf{h}_t$ & The hidden state vector representing the cognitive state at time $t$. & Latent Space \\

$\alpha_{i,j}$ & Attention weight between token $i$ and token $j$. & Micro-mechanics \\

$D(t)$ & Contextual Information Density (CID) at sequence position $t$. & Information Theory \\

$\Delta \psi$ & \textbf{Cognitive Drift}, measuring the divergence between agent perceptions. & Synchronization \\

$\Gamma$ & Sync Stability Index, the probability of systemic coherence. & Evaluation \\

$\tau$ & Logical Time, defined by semantic transitions rather than wall-clock time. & Temporal Logic \\
\bottomrule
\end{tabular}
\end{table}

\subsection{Formal Derivation of Semantic Slicing}
In Section 3, we introduced Semantic Slicing as a boundary detection problem. Here, we derive the transition criteria based on Attention Entropy.
\\Given a sequence of tokens $X = \{x_1, x_2, ..., x_n\}$, the attention mechanism computes a distribution $P_i$ for each token $x_i$ over the preceding context. The Contextual Information Density (CID), $D(t)$, is derived from the negative normalized entropy of this distribution:
\begin{equation}
H(P_t) = - \sum_{j=1}^{t} \alpha_{t,j}\log \alpha_{t,j}
\end{equation}

\begin{equation}
D(t)=1-\frac{H(P_t)}{\log(t)}
\end{equation}
The Boundary Criterion:
\\A Semantic Slice boundary $\partial \sigma$ is identified when the first-order derivative of $D(t)$ with respect to the sequence position exceeds a dynamic threshold $\epsilon$:
\begin{equation}
\frac{\partial D(t)}{\partial t} > \epsilon \;\Rightarrow\; t \in \partial \sigma
\end{equation}
This derivation proves that slices are not arbitrary partitions but represent phase transitions in information density, where the model shifts from "intra-concept" processing to "inter-concept" transition.

\subsection{Mathematical Model of Cognitive Drift (\texorpdfstring{$\Delta \psi$}{Delta psi})}
In Section 4, we posited that asynchronous execution leads to Cognitive Drift. We formalize this using a State-Space Model. Let $\Phi(\sigma, \theta)$ be the mapping of a semantic slice into a latent cognitive state $\mathbf{h}$.
\\For two agents $A$ and $B$, the drift at logical time $\tau$ is the cumulative divergence of their state trajectories:
\begin{equation}
h_A(\tau)=\Phi_A(\sigma_{\tau},\theta_A),\quad
h_B(\tau)=\Phi_B(\sigma_{\tau},\theta_B)
\end{equation}
The instantaneous drift $\delta(\tau)$ is defined by the Euclidean distance in the latent manifold $\mathcal{M}$:
\begin{equation}
\delta(\tau)=
\sqrt{
\sum_{k=1}^{d}
\left(h_{A,k}-h_{B,k}\right)^2
}
\end{equation}
The Total Cognitive Drift $\Delta \psi$ over an interaction interval $[0, T]$ is the integral:
\begin{equation}
\Delta \psi =
\int_{0}^{T}
e^{-\lambda (T-\tau)} \,\delta(\tau)\, d\tau
\end{equation}
where $\lambda$ is a Decay Constant representing the system's "forgetting factor." This integral demonstrates that without periodic Cognitive Sync Pulses (CSP) to reset $\delta(\tau) \to 0$, the system will inevitably cross the Entropy Barrier, leading to catastrophic decoherence.
\subsection{Derivation of the Sync Stability Index ($\Gamma$)} 
The stability of AgentOS is probabilistic. We define $\Gamma$ as the probability that the drift $\Delta \psi$ remains bounded by a task-specific coherence threshold $\epsilon_{max}$:
\begin{equation}
\Gamma = P\!\left(
\sup_{t \in [0,T]} \Delta \psi(t) < \epsilon_{\max}
\right)
\end{equation}
Assuming $\delta(\tau)$ follows a stochastic process (e.g., a Geometric Brownian Motion in the latent space), $\Gamma$ can be solved using the First-Passage Time theory. This provides the mathematical justification for our Advantageous-Timing Alignment—by synchronizing at points of minimum $\delta(\tau)$ (high-confidence windows), we maximize $\Gamma$ while minimizing the computational overhead of synchronization.

\section{AgentOS Reference Implementation Pseudocode} 
This appendix provides the high-level algorithmic logic required to implement the core subsystems of AgentOS. These pseudocodes abstract away hardware-specific details to focus on the semantic-level operations.
\subsection{The Semantic Paging and Eviction Logic} 
The Semantic Memory Management Unit (S-MMU) must handle the movement of context between the limited L1 (Attention Window) and the high-capacity L2 (Semantic RAM). Unlike classical OS paging, the eviction priority is determined by a Semantic Importance Score ($\mathcal{I}$), which is derived from the attention gradients calculated in Section 3.

\begin{algorithm}[H]
\caption{S-MMU Context Paging and Eviction (LRU-Semantic)}
\label{alg:smmu-paging}
\begin{algorithmic}[1]
\State \textbf{Input:} New Incoming Slice $\sigma_{in}$, Active Window $\mathcal{C}_{L1}$, Memory Limit $K$
\State \textbf{Output:} Synchronized L1 Context State

\Function{ManageMemory}{$\sigma_{in}$}
\While{$CurrentSize(\mathcal{C}_{L1}) + Size(\sigma_{in}) > K$}
\State $\sigma_{victim} \gets \arg\min_{\sigma \in \mathcal{C}_{L1}} \mathcal{I}(\sigma)$
\State \textbf{SAVE} $\sigma_{victim}$ to $L2\_Semantic\_RAM$
\State \textbf{UPDATE} Semantic Page Table (SPT) for $\sigma_{victim}$ with $Status = Paged\_Out$
\State \textbf{EVICT} $\sigma_{victim}$ from $\mathcal{C}_{L1}$
\EndWhile
\State \textbf{INSERT} $\sigma_{in} \to \mathcal{C}_{L1}$
\State \textbf{UPDATE} SPT for $\sigma_{in}$ with $Status = Active$
\State \Return $\mathcal{C}_{L1}$
\EndFunction

\end{algorithmic}
\end{algorithm}

\subsection{The Cognitive Sync Pulse (CSP) Orchestration} 
This algorithm defines how AgentOS maintains coherence across multiple independent Reasoning Kernels (RKs). The Sync Pulse acts as a global barrier that forces perception alignment when cognitive drift exceeds a safety threshold.

\begin{algorithm}[t]
\caption{Cognitive Sync Pulse and Multi-Agent Alignment}
\label{alg:csp-sync}
\begin{algorithmic}[1]

\State \textbf{Prerequisite:} Drift $\Delta \psi_i > \epsilon$ detected by the System Monitor

\Procedure{ExecuteSyncPulse}{$\{Agent_1, \dots, Agent_n\}$}

\State \textbf{STEP 1: SUSPEND} all Reasoning Threads across the cluster
\State \textbf{STEP 2: CAPTURE} the current Hidden State $\mathbf{h}_i$ and active slice $\sigma_i$ from each agent

\State \textbf{STEP 3: RESOLVE CONFLICTS}
\For{each $Slice\_Group$ covering the same logical time $\tau$}
\State $\sigma_{unified} \gets \text{AggregateSemanticSlices}(\{\sigma_1, \dots, \sigma_n\})$
\State $\mathbf{h}_{unified} \gets \text{AlignLatentStates}(\{\mathbf{h}_1, \dots, \mathbf{h}_n\})$
\EndFor

\State \textbf{STEP 4: REBROADCAST} $\sigma_{unified}$ and $\mathbf{h}_{unified}$ to all agent L1 caches
\State \textbf{STEP 5: RESET} all Drift Meters $\Delta \psi_i \gets 0$
\State \textbf{STEP 6: RESUME} all Reasoning Threads

\EndProcedure

\end{algorithmic}
\end{algorithm}

\subsection{The Reasoning Interrupt Vector Table} 
To formalize I/O, we define a standard Interrupt Vector Table (IVT). This allows the OS to handle external tool calls as first-class system events.

\begin{table}[H]
\centering
\caption{AgentOS Interrupt Vector Table (Standard Implementation)}
\label{tab:B1}
\begin{tabular}{llll}
\toprule
\textbf{Interrupt ID} & \textbf{Signal Name} & \textbf{Description} & \textbf{Priority} \\
\midrule
0x01 & SIG\_TOOL\_INVOKE & Reasoning Kernel requests an external API/Tool call. & High \\
0x02 & SIG\_CONTEXT\_FULL & L1 Cache has reached its attention capacity. & Medium \\
0x03 & SIG\_SYNC\_DRIFT & Cognitive Drift $\Delta \psi$ has exceeded threshold $\epsilon$. & High \\
0x04 & SIG\_PERCEPTION\_ERR & Output of tool does not match the current semantic schema. & Critical \\
\bottomrule
\end{tabular}
\end{table}

\section{Comparative Taxonomy of Autonomous Frameworks} 
To contextualize the architectural advancements of AgentOS, this appendix provides a detailed comparison with existing state-of-the-art (SOTA) agentic frameworks. We evaluate these systems based on their underlying system abstractions rather than their application-level performance.
\subsection{Comparative Matrix} 
Table C1 summarizes the structural differences between traditional "wrapper-based" frameworks and the system-level orchestration of AgentOS.

\begin{table}[H]
\centering
\caption{Architectural Taxonomy of LLM Agent Frameworks}
\label{tab:C1}
\resizebox{\linewidth}{!}{%
\begin{tabular}{lccccc}
\toprule
\textbf{Feature} & \textbf{AutoGen} & \textbf{MemGPT} & \textbf{BabyAGI} & \textbf{AIOS} & \textbf{AgentOS (Ours)} \\
\midrule
\textbf{Memory Model}
& Linear/Flat
& Hierarchical (Virtual)
& Flat/Queue-based
& Primitive Paging
& \textbf{Addressable Semantic Space} \\

\textbf{Addressing Unit}
& Token-based
& Page-based
& Task-based
& Token-block
& \textbf{Semantic Slicing ($\sigma$)} \\

\textbf{Scheduling}
& Sequential/Conversation
& Event-driven
& Priority Queue
& Round-robin
& \textbf{Cognitive Bandwidth Scheduling} \\

\textbf{Sync Mechanism}
& Turn-taking
& None
& None
& Basic Locking
& \textbf{Cognitive Sync Pulses (CSP)} \\

\textbf{I/O Abstraction}
& Function Call
& Tool Wrapper
& API Call
& System Call
& \textbf{Reasoning Interrupt Cycle (RIC)} \\

\textbf{Drift Management}
& Manual/Prompting
& None
& None
& None
& \textbf{Perception Alignment Protocol} \\
\bottomrule
\end{tabular}%
}
\end{table}

\subsection{Analysis of Architectural Paradigms} 
\subsubsection{From "Turn-taking" to "Sync Pulses"} 
Traditional frameworks like AutoGen manage multi-agent interaction through a conversational turn-taking paradigm. While intuitive, this approach is inherently synchronous and fails when agents must operate across disparate temporal scales. AgentOS moves beyond this by implementing Cognitive Sync Pulses, allowing for asynchronous execution with periodic deterministic re-alignment.

\subsubsection{Beyond Keyword Retrieval (S-MMU vs. RAG)} 
While MemGPT pioneered hierarchical memory, its retrieval mechanism often relies on traditional indexing. AgentOS introduces the S-MMU, which performs Semantic Paging. By utilizing attention-derived importance scores ($\mathcal{I}$), the system ensures that the most "cognitively dense" information remains in the L1 window, effectively mitigating the "lost-in-the-middle" effect observed in flat-context systems.

\subsubsection{The Evolution of Tool-Use} 
Current frameworks treat tool-use as an external library call. AgentOS formalizes this through the Reasoning Interrupt Cycle (RIC). By treating tools as hardware peripherals with associated interrupt vectors, the OS provides a resilient environment where tool errors or latencies do not crash the entire reasoning thread, but are instead handled via Perception Alignment.

\subsection{Summary of the Paradigm Shift} 
The transition from existing frameworks to AgentOS represents a shift from Application-logic to System-logic. As shown in Table C1, AgentOS is the first architecture to provide a mathematically grounded solution for Cognitive Drift and Semantic Addressing, providing a stable substrate for the emergence of high-order intelligence.

\end{document}